\documentclass{bmvc2k}

\usepackage{amsmath,amsfonts,bm}

\def\eqref#1{equation~\ref{#1}}

\def\1{\bm{1}}

\DeclareMathAlphabet{\mathsfit}{\encodingdefault}{\sfdefault}{m}{sl}
\SetMathAlphabet{\mathsfit}{bold}{\encodingdefault}{\sfdefault}{bx}{n}

\usepackage{tikz}
\usepackage{graphicx}
\usepackage{colortbl}
\usepackage{booktabs} 
\usepackage{multirow}
\usepackage{listings}
\usepackage{xcolor}
\usepackage{floatrow}
\usepackage{soul}
\usepackage{bibunits} 
\usepackage{fontawesome}
\usepackage{float}
\usepackage{caption}
\usepackage{newfloat}
\DeclareFloatingEnvironment[fileext=lol]{listing}

\newcommand{\ccolorbox}[1]{%
  \textcolor{#1}{\raisebox{0.5ex}{\fcolorbox{black}{#1}{\rule{0pt}{0.2ex}\rule{0.2ex}{0pt}}}}%
}

\newcommand{\hlorange}[1]{\sethlcolor{lightorange}\hl{#1}}  

\newcommand{\fsplit}{\texttt{Split-A}\xspace}
\newcommand{\ssplit}{\texttt{Split-B}\xspace}
\newcommand{\rsplit}{\texttt{Split-C}\xspace}

\newcommand{\maptf}{$\text{mAP}_{25}$}
\newcommand{\mapft}{$\text{mAP}_{50}$}

\definecolor{lightblue}{HTML}{9AB2DE}
\definecolor{lightorange}{HTML}{F9D5BF}
\definecolor{lightgreen}{HTML}{CDE8BE}
\definecolor{lightgray}{HTML}{D1D1D1}

\newcommand{\cpara}[1]{%
    \vspace{0.4\baselineskip}       
    \noindent\textbf{#1.} 
}

\usepackage{cleveref}
\crefname{figure}{Fig.}{Figs.}
\crefname{table}{Tab.}{Tabs.}
\crefname{algorithm}{Alg.}{Algs.}
\crefname{listing}{List.}{Lists.}
\crefname{equation}{Eq.}{Eqs.}
\crefname{section}{Sec.}{Secs.}

\definecolor{bmvcblue}{RGB}{29, 78, 137}  

\definecolor{bmvcblue}{rgb}{0, 0.1, 0.4} 

\let\titleold\title
\renewcommand{\title}[1]{\titleold{#1}\newcommand{\thetitle}{#1}}

\newcommand{\maketitlesupplementary}{
    \clearpage
    \begin{center}
        {\Large\bfseries\textcolor{bmvcblue}{\thetitle}}\\[0.5em]
        {\textcolor{bmvcblue}{Supplementary Material}}\\[1em]
    \end{center}
}

\definecolor{vscBlue}{RGB}{0,0,255}
\definecolor{vscGreen}{RGB}{0,128,0}
\definecolor{vscGray}{RGB}{128,128,128}
\definecolor{vscPurple}{RGB}{163,21,21}
\definecolor{vscOrange}{RGB}{255,128,0}
\definecolor{backcolour}{rgb}{1,1,1} 

\lstdefinestyle{mystyle}{
  backgroundcolor=\color{backcolour},   
  commentstyle=\color{vscGreen},        
  keywordstyle=\color{vscBlue}\bfseries, 
  numberstyle=\tiny\color{vscGray},     
  stringstyle=\color{vscPurple},        
  basicstyle=\ttfamily\footnotesize,    
  breakatwhitespace=false,              
  breaklines=true,                      
  captionpos=b,                         
  keepspaces=true,                      
  numbers=left,                         
  numbersep=5pt,                        
  showspaces=false,                     
  showstringspaces=false,
  showtabs=false,                       
  tabsize=2,
  xleftmargin=1.3em,   
  framexleftmargin=1.3em,  
}

\makeatletter
\let\lst@UseHook\relax
\makeatother

\title{CLIMB-3D: Class-Incremental Imbalanced 3D Instance Segmentation}

\addauthor{Vishal Thengane}{v.thengane@surrey.ac.uk}{1}
\addauthor{Jean Lahoud}{jean.lahoud@mbzuai.ac.ae}{2}
\addauthor{Hisham Cholakkal}{hisham.cholakkal@mbzuai.ac.ae}{2}
\addauthor{Rao Muhammad Anwer}{rao.anwer@mbzuai.ac.ae}{2}
\addauthor{Lu Yin}{l.yin@surrey.ac.uk}{1}
\addauthor{Xiatian Zhu}{xiatian.zhu@surrey.ac.uk }{1}
\addauthor{Salman Khan}{salman.khan@mbzuai.ac.ae}{2, 3}

\addinstitution{
 University of Surrey,\\
 Guildford, UK
}
\addinstitution{
 Mohamed bin Zayed University of Artificial Intelligence,\\
 Abu Dhabi, UAE
}
\addinstitution{
 Australian National University,\\
 Canberra, Australia
}

\runninghead{Thengane et al.}{CLIMB-3D}

\def\etal{\emph{et al}\bmvaOneDot}

\def\newsetup{CI-3DIS\xspace} 
\def\ourmethod{CLIMB-3D\xspace}
\def\ourmethodbf{\textbf{CLIMB-3D}\xspace}

\definecolor{cred}{RGB}{160,0,0}
\definecolor{cgreen}{RGB}{0,100,0}
\definecolor{cblue}{RGB}{0,51,179}

\begin{document}

\maketitle

\begin{abstract}
While 3D instance segmentation (3DIS) has advanced significantly, most existing methods 
assume that all object classes are known in advance and uniformly distributed. However, this assumption is unrealistic in dynamic, real-world environments where new classes emerge gradually and exhibit natural imbalance. 
Although some approaches address the emergence of new classes, they often overlook class imbalance, which leads to suboptimal performance, particularly on rare categories.
To tackle this, we propose \ourmethodbf, a unified framework for \textbf{CL}ass-incremental \textbf{Imb}alance-aware \textbf{3D}IS. 
Building upon established exemplar replay (ER) strategies, we show that ER alone is insufficient to achieve robust performance under memory constraints. 
To mitigate this, we introduce a novel pseudo-label generator (PLG) that extends supervision to previously learned categories by leveraging predictions from a frozen model trained on prior tasks. Despite its promise, PLG tends to be biased towards frequent classes. 
%
Therefore, we propose a class-balanced re-weighting (CBR) scheme that estimates object frequencies from pseudo-labels and dynamically adjusts training bias, without requiring access to past data.
We design and evaluate three incremental scenarios for 3DIS on the challenging ScanNet200 dataset and additionally validate our method for semantic segmentation on ScanNetV2.
Our approach achieves state-of-the-art results, surpassing prior work by up to 16.76\% mAP for instance segmentation and approximately 30\% mIoU for semantic segmentation, demonstrating strong generalisation across both frequent and rare classes.
Code is available at: \href{https://github.com/vgthengane/CLIMB3D}{\texttt{https://github.com/vgthengane/CLIMB3D}}.

\end{abstract}
    
\section{Introduction}
\label{sec:intro}

\begin{figure*}[!htb]
  \centering
  \includegraphics[width=\linewidth]{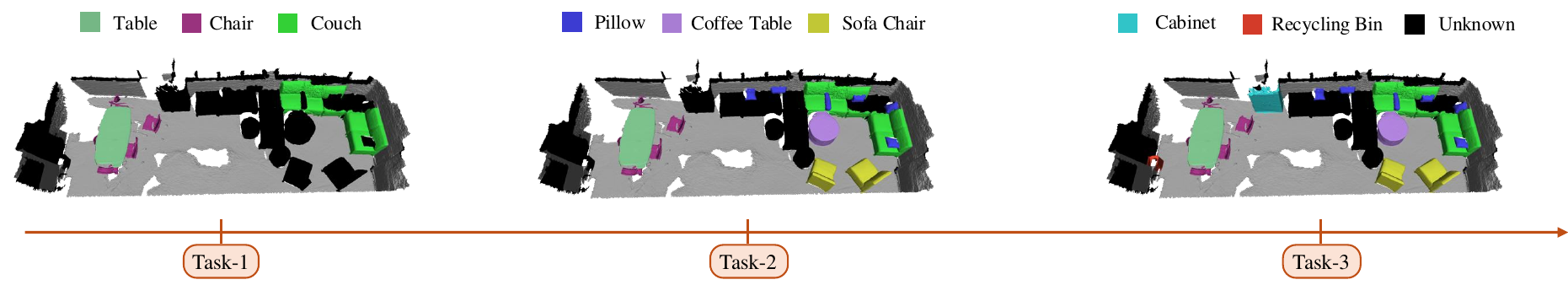}
  \caption{
  \textbf{Overview of the CI-3DIS setting.} New object categories are introduced incrementally with each task. After every phase, the model must recognise both newly added and previously learned classes. For instance, in Task~2, categories such as \texttt{\textit{Pillow}}, \texttt{\textit{Coffee Table}}, and \texttt{\textit{Sofa Chair}} are introduced, while the model is expected to retain recognition of earlier classes like \texttt{\textit{Table}}, \texttt{\textit{Chair}}, and \texttt{\textit{Couch}}.
  }
  \label{fig:cl_setup}
\end{figure*}

3D instance segmentation (3DIS) is a fundamental task in computer vision that involves identifying precise object boundaries and class labels in 3D space, with broad applications in graphics, robotics, and augmented reality \cite{boudjoghra20243d, lai2022stratified}. 
Traditional 3DIS methods, including top-down~\cite{wang2018sgpn, jia2021scaling, zhang2021point}, bottom-up~\cite{hou20193dsis, yang2019learning}, and transformer-based approaches~\cite{Schult23ICRA}, perform well under the assumption that all classes are available and balanced during training.
However, this assumption does not hold in real-world environments, where new object classes emerge over time and exhibit natural imbalance.

This gap motivates the need for Class-Incremental Learning (CIL) \cite{rebuffi2017icarl, de2021continual}, which supports learning new categories while retaining knowledge of previous ones~\cite{mccloskey1989catastrophic}. While CIL has seen success in 2D image tasks \cite{rebuffi2017icarl, li2017learning, aljundi2018memory, serra2018overcoming}, extensions to 3D point clouds remain limited, often focusing on object-level classification \cite{dong2021i3dol, liu2021l3doc, chowdhury2021learning}. 
%
Recent works on scene-level class-incremental 3DIS (CI-3DIS) \cite{boudjoghra20243d} and class-incremental semantic segmentation (CI-3DSS) \cite{Yang_2023_CVPR} show promise, but depend heavily on large exemplar  \cite{rolnick2019experience} and overlook class imbalance, limiting their practicality.

To address this, we propose \ourmethodbf, a unified framework for \textbf{CL}ass-incremental \textbf{IMB}alance-aware \textbf{3D}IS that jointly tackles catastrophic forgetting and class imbalance.
The framework begins with Exemplar Replay (ER), storing a small number of samples from past classes for later replay. However, this alone does not yield promising results under the strict memory constraints required in CI-3DIS.
To address this, we introduce a Pseudo-Label Generator (PLG), which uses a frozen model from previous task to generate supervision for earlier classes. However, we observed that PLG tends to favour frequent classes while ignoring under-represented ones. To mitigate this, we propose a Class-Balanced Re-weighting (CBR), which estimates class frequencies from the pseudo-labels to enhance the learning of rare classes without accessing past data. Together, these components form a practical and effective solution for real-world 3DIS.

To evaluate \ourmethod, we designed three benchmark scenarios for CI-3DIS based on the ScanNet200 dataset \cite{rozenberszki2022language}. These scenarios simulate incremental learning setup under natural class imbalance, where new classes emerge based on (A) object frequency, (B) semantic similarity, or (C) random grouping. 
Additionally, for comparison with existing methods, we evaluate our approach on CI-3DSS using the ScanNetV2 dataset \cite{dai2017scannet}. 
Experimental results show that our method significantly reduces forgetting and improves performance across both frequent and rare categories, outperforming previous methods in both settings. ~\cref{fig:cl_setup} illustrates the CI-3DIS setup.

In summary, our contributions are: (i) a novel problem setting of imbalanced class-incremental 3DIS with an effective method to balance learning and mitigate forgetting; (ii) three benchmarks modelling continual object emergence with natural imbalance; and (iii) strong experimental results, achieving up to 16.76\% mAP improvement over baselines.
\section{Related Work}
\label{sec:related_work}

\cpara{3D Instance Segmentation}
Various methods have been proposed for 3DIS. Grouping-based approaches adopt a bottom-up pipeline that learns latent embeddings for point clustering \cite{wang2018sgpn,lahoud20193d,jiang2020pointgroup,zhang2021point,chen2021hierarchical,han2020occuseg,he2021dyco3d,liang2021instance}. In contrast, proposal-based methods follow a top-down strategy by detecting 3D boxes and segmenting objects within them \cite{yang2019learning,hou20193dsis,liu2020learning,yi2019gspn,engelmann20203d}. 
Recently, transformer-based models \cite{vaswani2017attention} have also been applied to 3DIS and 3DSS \cite{Schult23ICRA,sun2022superpoint}, inspired by advances in 2D vision \cite{cheng2022masked,cheng2021per}. 
However, these approaches require full annotations for all classes and are not designed for progressive learning, where only new class annotations are available and previous data is inaccessible.
Another line of work aims to reduce annotation costs by proposing weakly supervised 3DIS methods based on sparse cues \cite{xie2020pointcontrast,hou2021exploring,chibane2022box2mask}. 
While effective with limited annotations, these methods assume a fixed set of classes and are susceptible to catastrophic forgetting in incremental settings.

\cpara{Incremental Learning}
Continual learning involves training models sequentially to mitigate catastrophic forgetting. One common strategy is model regularisation, which constrains parameter updates using techniques such as elastic weight consolidation or knowledge distillation \cite{kirkpatrick2017overcoming, aljundi2018memory, li2017learning, serra2018overcoming, hinton2015distilling}. Another approach is exemplar replay, where past data is stored or generated to rehearse older tasks while learning new ones \cite{rebuffi2017icarl, kamra2017deep, buzzega2020dark, cha2021co2l}. A third direction involves dynamically expanding the model architecture or using modular sub-networks to accommodate new tasks without interfering with old ones \cite{rusu2016progressive, li2019learn, zhao2022deep, wang2020learn, ke2020continual, rajasegaran2019random}.
Recent efforts in class-incremental 3D segmentation \cite{Yang_2023_CVPR, su2024balanced} have shown early promise but often rely on basic architectures and focus primarily on semantic segmentation. Other works \cite{kontogianni2024continual, boudjoghra20243d} explore continual learning in 3D settings, though they often depend on large memory buffers. In contrast, our work introduces a tailored framework for 3D instance segmentation that effectively transfers knowledge across tasks, even under class imbalance.

\cpara{Long-tailed Recognition}
Imbalanced class distributions lead to poor recognition of rare (long-tail) classes. To address this, re-sampling methods balance the data via input \cite{chawla2002smote, gupta2019lvis, park2022majority, shen2016relay, mahajan2018exploring} or feature space \cite{ando2017deep, chu2020feature, li2021metasaug}, avoiding naïve under-/over-sampling. Loss re-weighting offers another direction, using class-based \cite{khan2017cost, he2022relieving, cao2019learning, khan2019striking, cui2019class, wang2021adaptive} or per-example \cite{lin2017focal, ren2018learning, shu2019meta} adjustments to ensure fair contribution. Parameter regularisation improves generalisation through weight constraints \cite{alshammari2022long}, albeit requiring careful tuning. Other approaches leverage transfer learning \cite{yin2019feature, zhou2020bbn}, self-supervision \cite{yang2020rethinking, li2021self}, or contrastive learning \cite{khosla2020supervised, li2021self, zhu2022balanced} to improve rare-class representations. Long-tailed recognition is well-studied in 2D with large-scale datasets \cite{cui2019class, van2018inaturalist, liu2019large}; in 3D, ScanNet200 \cite{rozenberszki2022language, dai2017scannet} enables similar efforts. Prior 3D work focused on re-weighting, re-sampling, and transfer learning \cite{rozenberszki2022language}, while regularisation remains underexplored. CeCo \cite{zhong2023understanding} balances class centres via auxiliary loss but overlooks sampling and augmentation. Lahoud et al. \cite{lahoud2024long} propose adaptive classifier regularisation for 3D segmentation, outperforming prior approaches without threshold tuning. However, neither method considers incremental learning. In contrast, we jointly address long-tail and continual 3D semantic segmentation.
\section{Methodology}


\begin{figure*}[t]
  \centering
  \includegraphics[width=\linewidth]{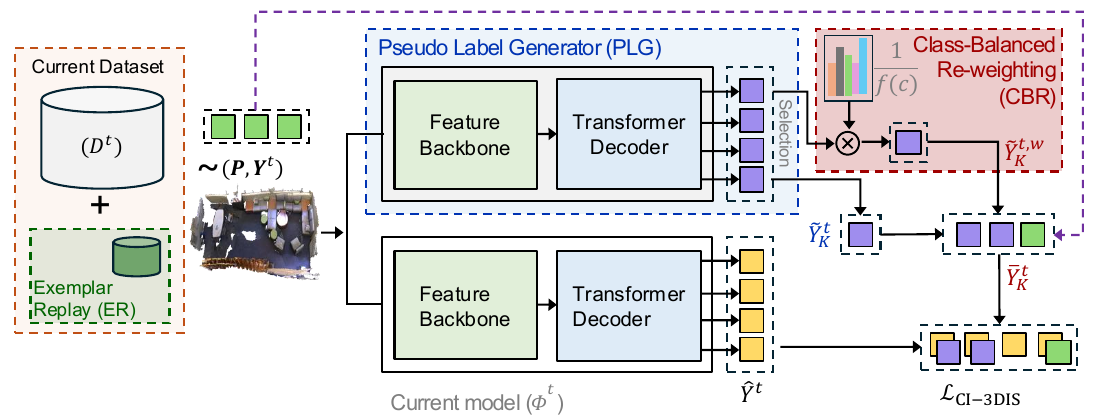}
  \caption{
  Overview of \ourmethodbf for \newsetup.
The model incrementally learns new classes across sequential phases. During task $t$, point clouds $\mathbf{P}$ and their corresponding labels $\mathbf{Y}^t$ are sampled from a combination of the current training dataset $D^t$ and {\color{cgreen}{Exemplar Replay (ER)}}, which maintains a small memory of past examples. These are then passed to the current model $\Phi^t$ to produce predicted labels $\mathbf{Y}^t$. The {\color{cblue}{Pseudo-Label Generator (PLG)}} selects the top-$K$ predictions from the previous model $\Phi^{t-1}$. These pseudo-labels are then weighted based on class frequency $f(c)$ using {\color{cred}{Class-Balanced Re-weighting (CBR)}}, and the top-$K$ re-weighted labels are selected to form the balanced pseudo-label set $\bar{\mathbf{Y}}^t$. This pseudo-label set is then concatenated with the ground-truth labels to form a final augmented supervision set $\overline{\mathbf{Y}}^t$ for task $t$, which is used to optimise the model $\Phi^t$ using \cref{eq:climb3d}.
  }
  \label{fig:climb3d}
\end{figure*}

\subsection{Problem Formulation}
\label{sec:formulation}

\cpara{3DIS} 
The objective of this task is to detect and segment each object instance within a point cloud. Formally, the training dataset is denoted as $\mathcal{D} = \{(\mathbf{P}_i, \mathbf{Y}_i)\}_{i=1}^N$, where $N$ is the number of training samples. Each input $\mathbf{P}_i \in \mathbb{R}^{M \times 6}$ is a coloured point cloud consisting of $M$ points, where each point is represented by its 3D coordinates and RGB values.
%
The corresponding annotation $\mathbf{Y}_i = \{(m_{i,j}, c_{i,j})\}_{j=1}^{J_i}$ contains $J_i$ object instances, where $m_{i,j} \in \{0,1\}^M$ is a binary mask indicating which points belong to the $j$-th instance, and $c_{i,j} \in \mathcal{C} = \{1, \dots, C\}$ is the semantic class label of that instance.
Given $\mathbf{P}_i$, the model $\Phi$ predicts instance-level outputs $\hat{\mathbf{Y}}_i = \{(\hat{m}_{i,j}, \hat{c}_{i,j})\}_{j=1}^{\hat{J}_i}$, where $\hat{m}_{i,j}$ and $\hat{c}_{i,j}$ denote the predicted mask and category label for the $j$-th instance, respectively. The number of predicted instances $\hat{J}_i$ 
varies depending on the model's inference.
The model is optimised using the following objective:
\begin{equation}
\mathcal{L}_{\text{3DIS}}(\mathcal{D}; \Phi) = 
\frac{1}{|\mathcal{D}|} \sum_{(\mathbf{P}, \mathbf{Y}) \in \mathcal{D}} 
\frac{1}{|\mathbf{Y}|} \sum_{(m_j, c_j) \in \mathbf{Y}} 
\left( \mathcal{L}_{\text{mask}}(m_j, \hat{m}_j) + \lambda \cdot \mathcal{L}_{\text{cls}}(c_j, \hat{c}_j) \right),
\label{eq:3diss}
\end{equation}
where %
$\mathcal{L}_{\text{cls}}$ and $\mathcal{L}_{\text{mask}}$ 
are the average classification and mask losses over all instances, 
and $\lambda$ controls their trade-off.

\cpara{CI-3DIS} %
Unlike conventional 3DIS, CI-3DIS involves sequential learning, where object categories are introduced incrementally over $T$ training tasks. 
Each task $t$ introduces a disjoint set of classes $\mathcal{C}^t$, with $\mathcal{C} = \bigcup_{t=1}^{T} \mathcal{C}^t$ and $\mathcal{C}^t \cap \mathcal{C}^{t'} = \emptyset$ for $t \neq t'$.
During each task $t$, the model receives a dataset $\mathcal{D}^t = \{(\mathbf{P}_i, \mathbf{Y}_i^t)\}_{i=1}^N$, 
where each coloured point cloud $\mathbf{P}_i \in \mathbb{R}^{M \times 6}$ contains $M$ points, and the corresponding annotation $\mathbf{Y}_i^t = \{(m_{i,j}^t, c_{i,j}^t)\}_{j=1}^{J_i^t}$ includes instance masks $m_{i,j}^t \in \{0,1\}^M$ and semantic class labels $c_{i,j}^t \in \mathcal{C}^t$.
%
%
The model $\Phi^t$, initialized from $\Phi^{t-1}$, is trained on $\mathcal{D}^t$ to predict instance-level outputs $\hat{\mathbf{Y}}_i^t = \{(\hat{m}_{i,j}^t, \hat{c}_{i,j}^t)\}_{j=1}^{\hat{J}_i^t}$, where each predicted semantic label $\hat{c}_{i,j}^t$ belongs to the cumulative set of all classes observed so far: $\bigcup_{k=1}^{t} \mathcal{C}^k$.
%
%
%
The key challenge in this setup is to learn new classes without forgetting those encountered in previous tasks, despite limited supervision and the absence of past labels.

\subsection{Method: CLIMB-3D}
\label{sec:climb3d}
\cpara{Overview}
As shown in \cref{fig:climb3d} and formulated in \cref{sec:formulation}, the proposed framework for \newsetup follows a \textit{phase-wise} training strategy. In each phase, the model is exposed to a carefully curated subset of the dataset, simulating real-world scenarios discussed in \cref{sec:scenarios}.
Training na\"ively on such phased data leads to \textit{catastrophic forgetting} \cite{mccloskey1989catastrophic}, where the model forgets 
prior knowledge when learning new tasks. 
To address this, we first incorporate \textit{Exemplar Replay} (\textbf{ER}) \cite{buzzega2020dark}, a strategy inspired by 2D incremental methods \cite{rebuffi2017icarl, li2017learning, aljundi2018memory} and recently extended to 3D settings \cite{boudjoghra20243d}, to mitigate forgetting by storing a small set of representative samples from earlier phases.
However, ER alone is insufficient to achieve robust performance. 
To address this, two additional components are introduced: a \textit{Pseudo-Label Generator} (\textbf{PLG}), which leverages a frozen model to generate supervision signals for previously seen classes, and a \textit{Class-Balanced Re-weighting} (\textbf{CBR}) module that compensates for class imbalance across phases.
Each of these components is described in detail below \footnote{For clarity, modifications introduced by each component (ER, PLG, and CBR) are highlighted in \textcolor{cgreen}{green}, \textcolor{cblue}{blue}, and \textcolor{cred}{red}, respectively.}.
%

\cpara{Exemplar Replay (ER)}
Inspired by Buzzega et al.~\cite{buzzega2020dark}, ER addresses the issue of catastrophic forgetting by allowing the model to retain a subset of data from earlier phases. During phase~$t$, the model is trained not only on the current task data $\mathcal{D}^t$, but also on a set of exemplars $\mathcal{E}^{1:t-1}$, collected from previous phases. 
These exemplars are accumulated incrementally: 
%
$\mathcal{E}^{1:t-1} = \mathcal{E}^1 \cup \dots \cup \mathcal{E}^{t-1}.$
Here, $\mathcal{E}^t$ denotes the exemplar subset stored after phase~$t$, and $|\mathcal{E}^t|$ denotes the exemplar replay size.
%
The combined replay dataset is denoted as:
${\color{cgreen}\mathcal{D}^t_{\text{ER}}} = \mathcal{D}^t \cup \mathcal{E}^{1:t-1}.$
Epoch training consists of two stages: first, the model is trained on the current data, $\mathcal{D}^t$; then it is trained using exemplars from $\mathcal{E}^{1:t-1}$. 
%
%
%

%
While previous \newsetup approaches rely on large exemplar sets to mitigate forgetting~\cite{boudjoghra20243d}, such strategies become impractical under memory constraints. Consequently, although ER aids knowledge retention, it is insufficient on its own for robust performance with a limited exemplar budget. 
Therefore, PLG and CBR are introduced to better preserve past representations and enhance generalisation across phases.

%
\cpara{Pseudo-Label Generator (PLG)} 
In the \newsetup setting, during phase $t$ (where $t > 1$), although ground-truth labels for previously seen classes are unavailable, the model from the previous phase $\Phi^{t-1}$, which preserves knowledge of past tasks, is retained.
We use this model to generate pseudo-labels for previously seen classes, thereby providing approximate supervision during training. This allows the current model $\Phi^t$, to retain prior knowledge and reduce forgetting, even without access to ground-truth annotations.

Given a point cloud and label pair $(\mathbf{P}, \mathbf{Y}^t)$, from $\mathcal{D}^t$, the previous model $\Phi^{t-1}$, generates pseudo-labels for the previously seen classes as $\hat{\mathbf{Y}}^{1:t-1} = \Phi^{t-1}(\mathbf{P})$, which we denote as $\tilde{\mathbf{Y}}^t$ for brevity. 
However, we observe that utilising all predictions from $\Phi^{t-1}$ introduces noisy or incorrect labels, which degrade overall performance.
To mitigate this, we select the top-$K$ most confident instance predictions, denoted as $\tilde{\mathbf{Y}}^t_K$, and concatenate them with the ground-truth labels of the current task to obtain the supervision set: ${\color{cblue}\bar{\mathbf{Y}}^t} = \mathbf{Y}^t \mathbin\Vert {\color{cblue}\tilde{\mathbf{Y}}^t_K}$. Following this,
%
%
the loss is computed over both the current task’s ground-truth labels and the top-$K$ confident pseudo-labels from earlier phases, enabling the model to retain prior knowledge while adapting to new classes.

\cpara{Class-Balanced Re-weighting (CBR)}
%
While ER and PLG contribute to preserving prior knowledge, we observe that the model's predictions are biased towards frequent classes, resulting in the forgetting of rare categories. 
%
%
%
This issue is further amplified by the top-$K$ pseudo-labels selected by PLG, which predominantly represent dominant classes in the data.
To address this, we propose a \textit{Class-Balanced Re-weighting} (CBR) scheme that compensates for class imbalance using object frequency statistics.
%
At task $t$, only the current dataset $\mathcal{D}^t$ and its label distribution $p^t(c)$ over classes $c \in \mathcal{C}^t$ are available.
%
Datasets and class distributions from earlier tasks $\{ \mathcal{D}^i, p^i(c) \}_{i=1}^{t-1}$ are no longer accessible, which makes it challenging to directly account for class imbalance across all previously seen tasks.
Therefore, we propose to leverage the pseudo-label predictions of the frozen model $\Phi^{t-1}$ as a proxy for the class distribution across previously learned categories.

During each training iteration, the previous model $\Phi^{t-1}$ is applied to the current input $\mathbf{P}$ to generate pseudo-labels $\tilde{\mathbf{Y}}^t$, 
from which we accumulate class-wise frequency statistics for previously learned classes. Let $\mathcal{C}^{1:t} = \bigcup_{i=1}^{t} \mathcal{C}^i$ represent the union of all classes seen up to task $t$. 
At the end of each epoch, we compute the overall class frequency $\mathbf{f} \in \mathbb{R}^{|\mathcal{C}^{1:t}|}$ by combining the pseudo-label distribution $\tilde{p}^t(c)$ derived from $\tilde{\mathbf{Y}}^t$ with the ground-truth label distribution $p^t(c)$ from the current task. 
%
%
The resulting frequency vector $\mathbf{f} = [f(c)]_{c \in \mathcal{C}^{1:t}}$ is then used to compute class-wise weights $\mathbf{w} = [w(c)]_{c \in \mathcal{C}^{1:t}}$, which guide balanced pseudo-label selection and help the model focus on rare classes.

In the next epoch, 
the model $\Phi^{t-1}$ produces 
soft pseudo-label predictions $\tilde{\mathbf{Y}}^t$
(as described in PLG)
, where each component $\tilde{\mathbf{Y}}^t[c]$ denotes the confidence score for class $c \in \mathcal{C}^{1:t-1}$. To promote balanced pseudo-label selection, 
class-wise re-weighting is applied using $w(c) = \frac{1}{f(c) + \epsilon}$, where $\epsilon$ is a small constant for numerical stability. 
The adjusted predictions are computed by applying the class-wise weights as:
%
$
\tilde{\mathbf{Y}}^{t,w}[c] = w(c) \cdot \tilde{\mathbf{Y}}^t[c], \quad \forall c \in \mathcal{C}^{1:t-1}.
$
%
Top-$K$ selection is then applied to both the original and the re-weighted scores, yielding two pseudo-label sets: $\tilde{\mathbf{Y}}^t_K$ and $\tilde{\mathbf{Y}}^{t,w}_K$, respectively. The final augmented target set, $\overline{\mathbf{Y}}^t$, is constructed by concatenating the ground-truth labels $\mathbf{Y}^t$ with both sets of pseudo-labels:
\begin{equation}
  {\color{cred}\overline{\mathbf{Y}}^t} = \mathbf{Y}^t \mathbin\Vert {\color{cblue}\tilde{\mathbf{Y}}^t_K} \mathbin\Vert {\color{cred}\tilde{\mathbf{Y}}^{t,w}_K}. 
  \label{eq:fin_targets}
\end{equation}
%
%

Although the above re-weighting mitigates bias from prior tasks, it does not address class imbalance within the current task $t$. To resolve this, 
the class-balanced weights $w_c$ are extended to incorporate statistics from both previously seen and current classes, i.e., over $\mathcal{C}^{1:t}$ rather than only $\mathcal{C}^{1:t-1}$.
The updated weights, denoted as
$w_c'$, are
used to re-weights both pseudo-label selection and classification loss, ensuring balanced supervision across all seen categories, including rare ones. This unified strategy reduces bias and enhances performance
across phases.
%
The pseudo-code for constructing the supervision set in \cref{eq:fin_targets}, which incorporates both PLG and CBR components, is provided in Appendix~{\color{red}A}.
%

\cpara{Final Objective} 
The final training objective of \ourmethodbf, incorporating ER, PLG, and CBR, is formulated in \cref{eq:3diss} as:
\begin{equation}
%
%
\mathcal{L}({\color{cgreen}\mathcal{D}^t_{\text{ER}}}; \Phi^t) = 
\frac{1}{|{\color{cgreen}\mathcal{D}^t_{\text{ER}}}|} \sum_{(\mathbf{P}, \mathbf{Y}^t) \in {\color{cgreen}\mathcal{D}^t_{\text{ER}}}} 
\frac{1}{|{\color{cred}\overline{\mathbf{Y}}^t}|} \sum_{({\color{cred}\overline{m}_j^t}, {\color{cred}\overline{c}_j^t}) \in {\color{cred}\overline{\mathbf{Y}}^t}} 
\left( \mathcal{L}_{\text{mask}}({\color{cred}\overline{m}_j^t}, \hat{m}_j^t) + {\color{cred}\mathbf{w}''_c} \cdot \mathcal{L}_{\text{cls}}({\color{cred}\overline{c}_j^t}, \hat{c}_j^t) \right),
\label{eq:climb3d}
\end{equation}
%
where ${\color{cred}\mathbf{w}''_c} = {\color{cred}\mathbf{w}'_c} \cdot \lambda_\mathrm{cls}$ denotes the scaled weight vector, and $\lambda_\mathrm{cls}$ is a hyperparameter for balancing the loss terms. 
%
%

\subsection{Benchmarking Incremental Scenarios}
\label{sec:scenarios}


While \newsetup methods offer a wide range of practical applications, they frequently rely on the assumption of uniform sample distribution, which rarely holds in real-world settings. In practice, the number of object categories, denoted by $\mathcal{C}$, is often large and characterised by substantial variability in category frequency, shape, structure, and size. To address these challenges, we propose three incremental learning scenarios, each designed to capture a different facet of real-world complexity. The design of these scenarios is detailed below; for further information and illustrations, refer to Appendix {\color{red}E} of the supplementary material.

\cpara{\fsplit: Frequency Scenarios}
This scenario acknowledges that datasets are often labelled based on the frequency of categories. To accommodate this, we propose a split where the model 
learns from the most frequent categories and subsequently incorporates the less frequent ones in later stages. By prioritising the training of frequently occurring categories, the model can establish a strong foundation before expanding 
to handle rare categories.

\cpara{\ssplit: Semantic Scenarios}
Beyond frequency, semantic similarity is crucial in real-world deployments. While objects often share visual or functional traits, models may encounter semantically different categories in new environments. To simulate this challenge, we introduce the \ssplit scenario. Here, categories are grouped based on their semantic relationships, and the model is incrementally trained on one semantic group at a time. This setup encourages the model to generalise across semantically similar categories and adapt more effectively when exposed to new ones. Unlike the \fsplit scenario, which organises learning based on category frequency, the \ssplit scenario may contain both frequent and infrequent categories within a single task, focusing on semantic continuity and transfer.

\cpara{\rsplit: Random Scenarios}
In some cases, data labelling is driven by the availability of objects rather than predefined criteria. To capture this, we introduce the \rsplit scenario, which represents a fully random setting where any class can appear in any task, resulting in varying levels of class imbalance. By exposing the model to such diverse and unpredictable distributions, we aim to improve its robustness in real-world situations where labelled data availability is inconsistent.

These incremental scenarios are designed to provide a more realistic representation of object distributions, frequencies, and dynamics encountered in the real world.

\section{Experiments}

\subsection{Setup}

\cpara{Datasets} 
We evaluate \ourmethodbf on ScanNet200 \cite{rozenberszki2022language}, which comprises 200 object categories and exhibits significant class imbalance, which makes it ideal for simulating and assessing real-world scenarios. 
In addition, we compare our approach against existing incremental learning methods using ScanNetV2 \cite{dai2017scannet} in the 3DSS setting. For this evaluation, we follow the standard training and validation splits defined in prior works \cite{Yang_2023_CVPR} to ensure consistency with  
existing methods.

\cpara{Evaluation Metrics}
We evaluate our method using \textit{Mean Average Precision} (mAP), a standard metric for 3DIS which provides a comprehensive measure of segmentation quality by accounting for both precision and recall.  
For comparison with existing 3DSS methods, we report the \textit{mean Intersection over Union} (mIoU), which quantifies the overlap between predicted and ground truth segments.
%
To assess the model’s ability to mitigate catastrophic forgetting in incremental settings, we use the \textit{Forgetting Percentage Points} (FPP) metric, as defined in \cite{liu2023continual}. FPP captures performance degradation by measuring the accuracy drop on the initially seen categories between the first and final training phases.

Detailed descriptions of the incremental scenarios (\fsplit, \ssplit, \rsplit) and implementation are provided in Appendix
\textcolor{red}{B}.

\subsection{Results and Discussion}
\begin{table}[!tbh]
    \centering
    \caption{Comparison between the proposed method and the baseline in the 3DIS setting, evaluated using mAP$_{25}$, mAP$_{50}$, mAP, and FPP after training across all phase.
    }
    \footnotesize
    \setlength{\tabcolsep}{0.5em}
    \begin{tabular}{llccccc} 
    \toprule
        \multirow{2}{*}{\textbf{Scenarios}} & \multirow{2}{*}{\textbf{Methods}} & \multicolumn{3}{c}{\textbf{Average Precision} $\uparrow$} & \multicolumn{2}{c}{\textbf{FPP} $\downarrow$} \\ 
        \cmidrule(lr){3-5} \cmidrule(lr){6-7}
        & & \textbf{mAP$_{25}$} & \textbf{mAP$_{50}$} & \textbf{mAP} & \textbf{mAP$_{25}$} & \textbf{mAP$_{50}$} \\   
        \midrule
         & Baseline & 16.46 & 14.29 & 10.44 & 51.30 & 46.82  \\
        \rowcolor{lightorange}
        \multirow{-2}{*}{\cellcolor{white} \fsplit}&  \ourmethodbf (Ours) & \textbf{35.69} & \textbf{31.05} & \textbf{22.72} & \textbf{3.44} & \textbf{2.63} \\  
        \midrule
         & Baseline & 17.22 &15.07 & 10.93 & 46.27 & 42.1 \\
        \rowcolor{lightorange}
        \multirow{-2}{*}{\cellcolor{white} \ssplit} &  \ourmethodbf (Ours) & \textbf{35.48} & \textbf{31.56} & \textbf{23.69} & \textbf{8.00} & \textbf{5.51} \\ 
        \midrule
         & Baseline & 25.65 & 21.08 & 14.85 & 31.68 & 28.84 \\
        \rowcolor{lightorange}
        \multirow{-2}{*}{\cellcolor{white} \rsplit} &  \ourmethodbf (Ours) & \textbf{31.59} & \textbf{26.78} & \textbf{18.93} & \textbf{9.10} & \textbf{7.89} \\ 
    \bottomrule
    \end{tabular}
    \label{tab:er_comparison}
\end{table}

To evaluate our proposed approach, we construct a baseline using ER~\cite{buzzega2020dark} for the 3DIS setting, as no existing incremental baselines are available for this task. 
In contrast, for the 3DSS setting, where incremental baselines do exist, we compare our method against three prior approaches~\cite{serra2018overcoming, Yang_2023_CVPR, li2017learning}.

%
\cpara{Results on CI-3DIS}
\cref{tab:er_comparison} compares \ourmethod against the ER baseline across the three CI-3DIS scenarios, evaluated after all phases using \maptf, \mapft, overall mAP, and FPP. 
In the \fsplit scenario, characterised by significant distribution shifts, our method achieves gains of +19.23\%, +16.76\%, and +12.28\% in \maptf, \mapft, and overall mAP, respectively, while drastically reducing forgetting with FPP scores of 3.44\% (\maptf) and 2.63\% (\mapft), both over 45 points lower than the baseline.
Under the \ssplit scenario, with semantically related new categories, improvements of +18.26\%, +16.49\%, and +12.76\% are observed alongside a reduction in forgetting to 8.00\% and 5.51\% from baseline levels above 42\%. 
For the \rsplit scenario, featuring increasing geometric complexity, \ourmethod maintains solid gains of +5.94\%, +5.70\%, and +4.08\% across mAP metrics, with forgetting reduced by over 22 points. 
Overall, these results demonstrate robust forward transfer, minimal forgetting, and stable learning, highlighting the method’s effectiveness for scalable continual 3D semantic understanding.

\cpara{Results on CI-3DSS}
\begin{table}[t]
    \centering
    \caption{Comparison 
    between the proposed method and 
    existing 
    baselines 
    in the 3DSS setting on ScanNetV2 \cite{dai2017scannet}, evaluated using mIoU.}
    \setlength{\tabcolsep}{0.95em}
    \footnotesize
    \begin{tabular}{lrrr}
    \toprule
    \textbf{Methods} & \textbf{Phase=1} & \textbf{Phase=2} & \textbf{All} \\ 
    \midrule
    EWC \cite{serra2018overcoming} & 17.75 & 13.22 & 16.62 \\ 
    LwF \cite{li2017learning} & 30.38 & 13.37 & 26.13 \\ 
    Yang \etal \cite{Yang_2023_CVPR} & 34.16 & 13.43 & 28.98 \\ 
    \midrule
    \rowcolor{lightorange} \ourmethodbf (Ours) & \textbf{69.39} & \textbf{32.56} & \textbf{59.38} \\
    \bottomrule
    \end{tabular}
    \label{tab:sseg_compare}
\end{table}
%
Although our primary focus is on CI-3DIS, we also evaluate our method under the CI-3DSS setting to enable comparisons with existing approaches. To do so, we adapt our predictions to assign each point the label corresponding to the highest-confidence mask and exclude background classes (e.g., floor and wall), as these are not part of the semantic segmentation targets.
\cref{tab:sseg_compare} presents a comparison between our method and existing baselines on the ScanNet V2 dataset \cite{dai2017scannet}, evaluated using mIoU across two training phases and overall. Although originally designed for instance segmentation, our method generalises effectively to semantic segmentation, achieving substantial gains of +35.23\% mIoU in Phase 1 and +19.1\% in Phase 2. Overall, it achieves 59.38\% mIoU, significantly higher than the $\sim$30\% mIoU of prior methods,  highlighting its robustness and transferability.

Refer to Appendix~\textcolor{red}{C}, \textcolor{red}{D}, and \textcolor{red}{F} of the supplementary material for detailed analyses of per-phase performance, rare-class evaluation, and qualitative result comparisons, respectively.
\subsection{Ablation}

\begin{table}[tbh]
    \centering
    \caption{Ablation study illustrating the impact of 
    each component 
    in a three-phase setup. Each split (\textbf{s}) and corresponding number indicates data introduced at that phase.
    Best results are highlighted in \textbf{bold}.}
    \footnotesize
    \setlength{\tabcolsep}{4pt}
    \begin{tabular}{llr|rrr|rrrr|r} 
    \toprule
        \multirow{2}{*}{\textbf{Row}} & \multirow{2}{*}{\textbf{Modules}} &  \multicolumn{1}{c}{\textbf{p=1} $\uparrow$} & \multicolumn{3}{c}{\textbf{p=2} $\uparrow$ } & \multicolumn{4}{c}{\textbf{p=3} $\uparrow$} & \multirow{2}{*}{\textbf{FPP} $\downarrow$}  \\
        \cmidrule(lr){3-3} \cmidrule(lr){4-6} \cmidrule(lr){7-10}
        & & \textbf{s1}  & \textbf{s1} &  \textbf{s2} & \textbf{All} & \textbf{s1} &  \textbf{s2} &  \textbf{s3} & \textbf{All} & \\
        
        \midrule
        \rowcolor{lightgray!50} 1. & Oracle & - & - & - & - & 55.14 & 30.77 & 25.30 & 37.68 & - \\
        2. & Na\"ive & 56.82 & 0.00 & 28.09  & 14.15 & 0.00 & 0.00 & 19.67 & 5.80 & 56.82 \\
        3. & +~ER & 56.82 & 18.51 & 32.81  & 25.72 & 10.38 & 9.43 & 24.27 & 14.28 & 46.44 \\
        4. & +~PLG & 56.82 & 50.00 & \textbf{34.39} & 42.13 & 49.78 & 11.41 & 26.47 & 29.28 & 7.04 \\
        \rowcolor{lightorange} 5. & +~CBR & 56.82 & \textbf{54.67} & 33.75 & \textbf{44.13} & \textbf{54.19} & \textbf{12.02} & \textbf{26.55} & \textbf{31.05} & \textbf{2.63} \\
    \bottomrule
    \end{tabular}
    \label{tab:ablation}
\end{table}
We conduct an ablation study to evaluate the contribution of each component in our framework. As an upper bound, we report the performance of an \textit{Oracle} model, which is trained jointly on the full dataset. For the incremental setting, we follow the previously defined splits and first train the model naively across phases. We then incrementally add each module to isolate its impact. \cref{tab:ablation} presents results for the \fsplit scenario using both \mapft~ and FPP metrics.

\cpara{Na\"ive Training} When trained na\"ively without any dedicated modules, the model suffers from severe catastrophic forgetting, as evident in row 2. The model entirely forgets previously learned classes upon entering new phases, with performance dropping to zero for earlier splits and FPP rising to 56.82.

\cpara{Effect of ER} Adding exemplar replay (row 3) partially alleviates forgetting by maintaining a buffer of past examples. This yields notable gains for \textbf{s1} in Phase~2 (+18.51\%) and Phase~3 (+10.38\%), and also improves \textbf{s2} in Phase~3 (+9.43\%). However, the overall forgetting remains substantial (FPP: 46.44), showing ER alone is insufficient.

\cpara{Effect of PLG} The addition of the pseudo-label generator (PLG), which generates labels for previous classes by retaining a copy of the model from an earlier phase, facilitates knowledge retention and forward transfer. As shown in row 4, PLG significantly reduces forgetting and enhances performance on current tasks. For \textbf{s1}, it improves \mapft~by 31.49\% in Phase~2 and 39.40\% in Phase~3 over exemplar replay. Overall, PLG yields a 15.00\% increase in performance and reduces forgetting by 39.40\%.

\cpara{Effect of CBR} Finally, class-balanced re-weighting (CBR) mitigates class imbalance during both pseudo-labelling and current task learning by adjusting each class’s contribution based on its frequency. As shown in row 5, CBR enhances \textbf{s1} retention over PLG in Phases~2 and 3 (+4.67\% and +4.41\%), and further improves performance on \textbf{s2} and \textbf{s3}. It also achieves the lowest FPP of 2.63, reflecting strong forgetting mitigation. Overall, CBR offers the best trade-off between retaining past knowledge and acquiring new tasks, outperforming PLG and ER by 4.41\% and 43.81\%, respectively.

\subsection{Limitations \& Future Work}

While \ourmethod achieves strong performance across CI-3DIS and CI-3DSS settings, several limitations remain and warrant discussion. The experiments are currently limited to indoor datasets (ScanNet200 and ScanNetV2), therefore evaluating performance on outdoor scenes would help assess generalisability across diverse environments. The current setup also considers only three incremental phases, whereas real-world applications often involve longer sequences; incorporating such extended setups would better reflect practical challenges. Moreover, the model operates in a uni-modal setting. Integrating multi-modal cues, such as vision-language models, could further enhance performance, as suggested by recent 2D and 3D studies \cite{hong20233d, thengane2022clip, thengane2025foundational}.

\section{Conclusion}
This work addresses the challenge of catastrophic forgetting in class-incremental 3D instance segmentation by introducing a modular framework that integrates exemplar replay (ER), a pseudo-label generator (PLG), and class-balanced re-weighting (CBR). The proposed method incrementally adapts to new classes while preserving prior knowledge, all without requiring access to the full dataset. 
Extensive experiments on a three-phase benchmark validate the individual and combined effectiveness of each component: ER enables efficient memory-based retention; PLG facilitates knowledge preservation through pseudo-supervision; and CBR mitigates class imbalance to enhance learning stability. Together, these components significantly reduce forgetting and improve segmentation performance across all phases.
By achieving a strong balance between stability and plasticity, our approach advances continual 3D scene understanding and sets a new baseline for future research in this area.

\bibliography{bmvc_final}
\clearpage
\setcounter{page}{1}
\maketitlesupplementary

\appendix 
\renewcommand{\thesection}{Appendix \Alph{section}}

\section{Illustrating Incremental scenarios}
\label{app:scenarios}
\begin{figure}[h]
  \centering
  \includegraphics[width=\linewidth]{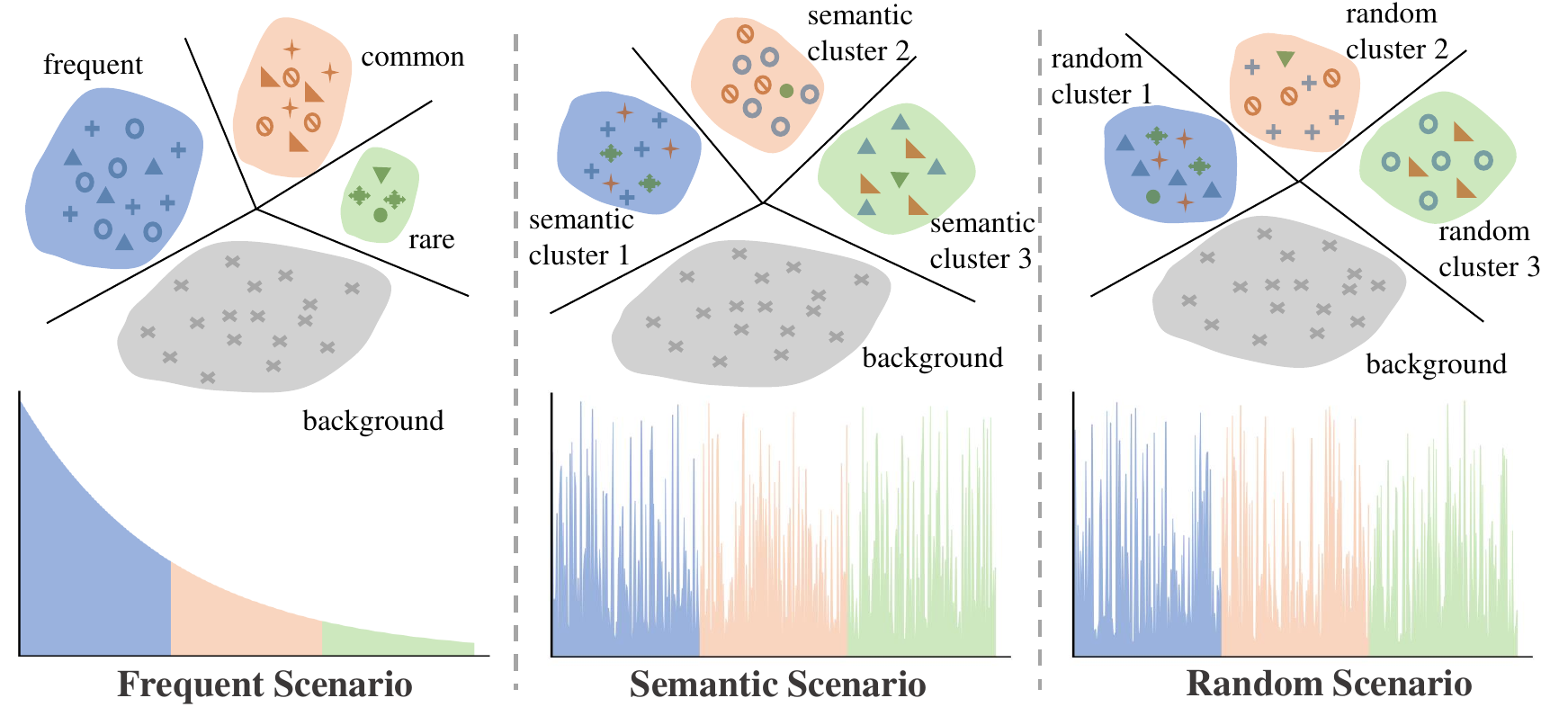}
  \caption{Tasks are grouped into incremental scenarios based on object frequency, semantic similarity, and random assignment. \ccolorbox{lightblue}{}, \ccolorbox{lightorange}{}, and \ccolorbox{lightgreen}{} denote different tasks; shapes indicate object categories; \ccolorbox{lightgray}{} marks the background. \textbf{Left:} Grouped by category frequency. \textbf{Middle:} Grouped by semantic similarity (e.g., similar shapes). \textbf{Right:} Randomly grouped, mixing semantic and frequency variations.}
  \label{fig:scenarios}
\end{figure}

\section{Per Phase Analysis}
\label{app:perphase}

\begin{table}[t]
    \centering
    \caption{Comparison of results in terms of mAP$_{50}$ with the proposed \ourmethod across three different scenarios. Each scenario is trained over three phases ($\mathtt{phase}=1,2,3$), introducing a single split \textbf{s} at a time. Results highlighted in \hlorange{orange} correspond to the proposed method, and the best results for each scenario are shown in \textbf{bold}.}
    \setlength{\tabcolsep}{4pt}
    \begin{tabular}{llc|ccc|ccccc} 
    \toprule
        \multirow{2}{*}{\textbf{Scenarios}} &  \multirow{2}{*}{\textbf{Methods}} & \multicolumn{1}{c}{\textbf{phase=1}} & \multicolumn{3}{c}{\textbf{phase=2}} & \multicolumn{4}{c}{\textbf{phase=3}}  \\ 
        \cmidrule(lr){3-3} \cmidrule(lr){4-6} \cmidrule(lr){7-10}
        & & \textbf{s1} & \textbf{s1} & \textbf{s2} & \textbf{All} & \textbf{s1} & \textbf{s2} & \textbf{s3} & \textbf{All} \\
        
        \midrule
         & Baseline & 56.82 & 18.51 & 32.81 & 25.72 & 10.38 & 9.43 & 24.27 & 14.28  \\
        \rowcolor{lightorange}
        \multirow{-2}{*}{\cellcolor{white} \fsplit} &  \ourmethodbf & 56.82 & \textbf{54.67} & \textbf{33.75} & \textbf{44.13} & \textbf{54.19} & \textbf{12.02} & \textbf{26.55} & \textbf{31.05}  \\
        \midrule
        & Baseline & 51.57 & 13.32 & \textbf{42.21} & 24.53 & 9.55 & 12.45 & \textbf{26.78} & 15.07 \\
        \rowcolor{lightorange}
        \multirow{-2}{*}{\cellcolor{white} \ssplit} &  \ourmethodbf & 51.57 & \textbf{46.74} & 37.45 & \textbf{43.13} & \textbf{46.06} & \textbf{15.95} & 26.68 & \textbf{31.56} \\ 
        \midrule
         & Baseline & 36.40 & 7.74 & \textbf{37.62} & 22.32 & 7.55 &  15.96 & \textbf{40.41} & 21.08 \\
        \rowcolor{lightorange}
        \multirow{-2}{*}{\cellcolor{white}  \rsplit} &  \ourmethodbf & 36.40 & \textbf{32.63} & 33.38 & \textbf{33.00} &
\textbf{28.51} & \textbf{17.11} & 34.64 & \textbf{26.78} \\ 
    \bottomrule
    \end{tabular}
    \label{tab:ph_comparison}
\end{table}

We extend the analysis from \cref{tab:er_comparison} to \cref{tab:ph_comparison} to highlight the impact of our proposed method on individual splits across various scenarios. The results clearly demonstrate that our model consistently retains knowledge of previous tasks better than the baseline. For \fsplit, our model shows improvement throughout the phase. In Phase 3 of (\textbf{s2}), although both the baseline and our method exhibit a performance drop, our method reduces forgetting significantly compared to the baseline.
The \ssplit scenario, while more complex than \fsplit, achieves comparable results due to semantic similarity among classes within the same task. In Phase 2, our model achieves overall all 43.13\% \mapft~compared to 24.53\% on baseline, a similar trend is observed in Phase 3, where our method not only consistently improves learning but also enhances retention of previous information. After all three tasks, our method achieves an overall performance of 31.56\% AP50, compared to 15.07\% for the baseline.
%
In the \rsplit scenario, the first-stage model struggles due to the increased complexity introduced by random grouping. In Phase 2, while the baseline focuses on learning the current task, it suffers from severe forgetting of prior knowledge. Conversely, our method balances new task learning with the retention of earlier information. By Phase 3, the model effectively consolidates \textbf{s1} and maintains strong performance across all task splits. Overall, our proposed method improves mAP by 5.6\%.

In this supplementary material, we first demonstrate the performance gains on rare classes achieved by incorporating the IC module in \ref{app:rare_eval}. Next, we provide detailed split information for all scenarios, based on class names, in \ref{app:split_fullview}. Finally, we present a qualitative comparison between the baseline method and our proposed approach in \ref{app:qual_results}.

\section{Evaluation on Rare Categories}
\label{app:rare_eval}
The proposed imbalance correction (IC) module, as detailed in Section 4.2, is designed to address the performance gap for rare classes. To assess its impact, we compare its performance with the framework which has exemplar replay (ER) and knowledge distillation (KD). Specifically, we focus on its ability to improve performance for rare classes, which the model encounters infrequently compared to more common classes.

\begin{table}[!ht]
    \centering
    \caption{Results for classes observed by the model 1–20 times during an epoch, evaluated on \fsplit~ for Phase 2, in terms of \mapft.}
    \small
    \resizebox{\columnwidth}{!}{%
    \setlength{\tabcolsep}{2pt}
    \begin{tabular}{lrrr}
    \toprule
    \rowcolor{lightorange}   \textbf{Classes} & \textbf{Seen Count} & \textbf{ER + PLG} & \textbf{ER + PLG + CBR} \\ 
    \midrule
        paper towel dispenser & 2 & 73.10 & 74.90 \\ 
        recycling bin & 3 & 55.80 & 60.50 \\ 
        ladder & 5 & 53.90 & 57.10 \\ 
        trash bin & 7 & 31.50 & 57.30 \\ 
        bulletin board & 8 & 23.30 & 38.20 \\ 
        shelf & 11 & 48.00 & 50.50 \\ 
        dresser & 12 & 44.00 & 55.80 \\ 
        copier & 12 & 93.30 & 94.50 \\ 
        object & 12 & 3.10 & 3.30 \\ 
        stairs & 13 & 51.70 & 67.70 \\ 
        bathtub & 16 & 80.30 & 86.60 \\ 
        oven & 16 & 1.50 & 3.30 \\ 
        divider & 18 & 36.40 & 45.00 \\ 
        column & 20 & 57.30 & 75.00 \\ 
    \midrule
        \rowcolor{lightgray} \textbf{Average} & - & 46.66 & \textbf{54.98} \\ 
    \bottomrule
    \end{tabular}
    }
    \label{tab:common_eval}
\end{table}

The results, shown in \Cref{tab:common_eval} and \Cref{tab:tail_eval}, correspond to evaluations on \fsplit~ for \textit{Phase 2} and \textit{Phase 3}, respectively. In \textit{Phase 2}, we evaluate classes seen 1–20 times per epoch, while \textit{Phase 3} targets even less frequent classes, with observations limited to 1–10 times per epoch. 

As illustrated in \Cref{tab:common_eval}, the IC module substantially improves performance on rare classes in terms of \mapft~ in Phase 2 of \fsplit. For instance, classes like \texttt{recycling bin} and \texttt{trash bin}, seen only 3 and 7 times, respectively, shows significant improvement when the IC module is applied. Overall, the IC module provides an average boost of 8.32\%, highlighting its effectiveness in mitigating class imbalance.

\begin{table}[!ht]
    \centering
    \caption{Results for classes observed by the model 1–10 times during an epoch, evaluated on \fsplit~ for Phase 3, in terms of \mapft.}
    \resizebox{\columnwidth}{!}{%
    \setlength{\tabcolsep}{10pt}
    \begin{tabular}{lrrr}
    \toprule
    \rowcolor{lightorange}   \textbf{Classes} & \textbf{Seen Count} & \textbf{ER+KD} & \textbf{ER+KD+IC} \\ 
    \midrule
        piano & 1 & 7.10 & 59.40 \\ 
        bucket & 1 & 21.10 & 31.50 \\ 
        laundry basket & 1 & 3.80 & 17.40 \\ 
        dresser & 2 & 55.00 & 55.40 \\ 
        paper towel dispenser & 2 & 32.50 & 35.50 \\ 
        cup & 2 & 24.70 & 30.30 \\ 
        bar & 2 & 35.40 & 39.50 \\ 
        divider & 2 & 28.60 & 42.40 \\ 
        case of water bottles & 2 & 0.00 & 1.70 \\ 
        shower & 3 & 0.00 & 45.50 \\ 
        mirror & 8 & 56.00 & 68.80 \\ 
        trash bin & 4 & 1.10 & 2.70 \\ 
        backpack & 5 & 74.50 & 76.70 \\ 
        copier & 5 & 94.00 & 96.80 \\ 
        bathroom counter & 3 & 3.90 & 20.30 \\ 
        ottoman & 4 & 32.60 & 36.20 \\ 
        storage bin & 3 & 5.10 & 10.50 \\ 
        dishwasher & 3 & 47.40 & 66.20 \\ 
        trash bin & 4 & 1.10 & 2.70 \\ 
        backpack & 5 & 74.50 & 76.70 \\ 
        copier & 5 & 94.00 & 96.80 \\ 
        sofa chair & 6 & 14.10 & 43.50 \\ 
        file cabinet & 6 & 49.20 & 57.60 \\ 
        tv stand & 7 & 67.70 & 68.60 \\ 
        mirror & 8 & 56.00 & 68.80 \\ 
        blackboard & 8 & 57.10 & 82.80 \\ 
        clothes dryer & 9 & 1.70 & 3.20 \\ 
        toaster & 9 & 0.10 & 25.90 \\ 
        wardrobe & 10 & 22.80 & 58.80 \\ 
        jacket & 10 & 1.20 & 4.10 \\ 
    \midrule
    \rowcolor{lightgray} \textbf{Average} & - & 32.08 & \textbf{44.21} \\ 
    \bottomrule
    \end{tabular}
    }
    \label{tab:tail_eval}
\end{table}
Similarly, \Cref{tab:tail_eval} presents results for \textit{Phase 3}, demonstrating significant gains for infrequent classes. For example, even though the classes such as \texttt{piano}, \texttt{bucket}, and \texttt{laundry basket} are observed only once, IC module improves the performance by 52.30\%, 10.40\%, and 13.60\%, respectively. The ER+KD module does not focus on rare classes like \texttt{shower} and \texttt{toaster} which results in low performance, but the IC module compensates for this imbalance by focusing on underrepresented categories. On average, the addition of the proposed IC module into the framework outperforms ER+KD by 12.13\%.

\begin{table*}[!ht]
    \centering
    \caption{Classes grouped by tasks for each proposed scenario on the ScanNet200 dataset labels. The three scenarios \fsplit, \fsplit, and \rsplit~ are each divided into three tasks: Task 1, Task 2, and Task 3.}
    \resizebox{\linewidth}{!}{%
    \setlength{\tabcolsep}{0.2em}
    \begin{tabular}{lll|lll|lll}
    \toprule
        \multicolumn{3}{c}{\fontfamily{lmtt}\fontseries{b}\selectfont{Split\_A}} & \multicolumn{3}{c}{\fontfamily{lmtt}\fontseries{b}\selectfont{Split\_B}} & \multicolumn{3}{c}{\fontfamily{lmtt}\fontseries{b}\selectfont{Split\_C}} \\ 
        \cmidrule(l{0.5cm}r{0.5cm}){1-3} \cmidrule(l{0.5cm}r{0.5cm}){4-6} \cmidrule(l{0.5cm}r{0.5cm}){7-9}
        \multicolumn{1}{l}{\textbf{Task 1}} & \multicolumn{1}{l}{\textbf{Task 2}} & \multicolumn{1}{l}{\textbf{Task 3}} & \multicolumn{1}{l}{\textbf{Task 1}} & \multicolumn{1}{l}{\textbf{Task 2}} & \multicolumn{1}{l}{\textbf{Task 3}} & \multicolumn{1}{l}{\textbf{Task 1}} & \multicolumn{1}{l}{\textbf{Task 2}} & \multicolumn{1}{l}{\textbf{Task 3}} \\ 
        \midrule
        chair & wall & pillow & tv stand & cushion & paper & broom & fan & rack \\ 
        table & floor & picture & curtain & end table & plate & towel & stove & music stand \\ 
        couch & door & book & blinds & dining table & soap dispenser & fireplace & tv & bed \\ 
        desk & cabinet & box & shower curtain & keyboard & bucket & blanket & dustpan & soap dish \\ 
        office chair & shelf & lamp & bookshelf & bag & clock & dining table & sink & closet door \\ 
        bed & window & towel & tv & toilet paper & guitar & shelf & toaster & basket \\ 
        sink & bookshelf & clothes & kitchen cabinet & printer & toilet paper holder & rail & doorframe & chair \\ 
        toilet & curtain & cushion & pillow & blanket & speaker & bathroom counter & wall & toilet paper \\ 
        monitor & kitchen cabinet & plant & lamp & microwave & cup & plunger & mattress & ball \\ 
        armchair & counter & bag & dresser & shoe & paper towel roll & bin & stand & monitor \\ 
        coffee table & ceiling & backpack & monitor & computer tower & bar & armchair & copier & bathroom cabinet \\ 
        refrigerator & whiteboard & toilet paper & object & bottle & toaster & trash bin & ironing board & shoe \\ 
        tv & shower curtain & blanket & ceiling & bin & ironing board & dishwasher & radiator & blackboard \\ 
        nightstand & closet & shoe & board & ottoman & soap dish & lamp & keyboard & vent \\ 
        dresser & computer tower & bottle & stove & bench & toilet paper dispenser & projector & toaster oven & bag \\ 
        stool & board & basket & closet wall & basket & fire extinguisher & potted plant & paper bag & paper \\ 
        bathtub & mirror & fan & couch & fan & ball & coat rack & structure & projector screen \\ 
        end table & shower & paper & office chair & laptop & hat & end table & picture & pillar \\ 
        dining table & blinds & person & kitchen counter & person & shower curtain rod & tissue box & purse & range hood \\ 
        keyboard & rack & plate & shower & paper towel dispenser & paper cutter & stairs & tray & coffee maker \\ 
        printer & blackboard & container & closet & oven & tray & fire extinguisher & couch & handicap bar \\ 
        tv stand & rail & soap dispenser & doorframe & rack & toaster oven & case of water bottles & telephone & pillow \\ 
        trash can & radiator & telephone & sofa chair & piano & mouse & water bottle & shower curtain rod & decoration \\ 
        stairs & wardrobe & bucket & mailbox & suitcase & toilet seat cover dispenser & ledge & trash can & printer \\ 
        microwave & column & clock & nightstand & rail & storage container & shower head & closet wall & object \\ 
        stove & ladder & stand & washing machine & container & scale & guitar case & cart & mirror \\ 
        bin & bathroom stall & light & picture & telephone & tissue box & kitchen cabinet & hat & ottoman \\ 
        ottoman & shower wall & pipe & book & stand & light switch & poster & paper cutter & water pitcher \\ 
        bench & mat & guitar & sink & light & crate & candle & storage organizer & refrigerator \\ 
        washing machine & windowsill & toilet paper holder & recycling bin & laundry basket & power outlet & bowl & vacuum cleaner & divider \\ 
        copier & bulletin board & speaker & table & pipe & sign & plate & mouse & toilet \\ 
        sofa chair & doorframe & bicycle & backpack & seat & projector & person & paper towel roll & washing machine \\ 
        file cabinet & shower curtain rod & cup & shower wall & column & candle & storage bin & laundry detergent & mat \\ 
        laptop & paper cutter & jacket & toilet & bicycle & plunger & microwave & calendar & scale \\ 
        paper towel dispenser & shower door & paper towel roll & copier & ladder & stuffed animal & office chair & wardrobe & dresser \\ 
        oven & pillar & machine & counter & jacket & headphones & clothes dryer & whiteboard & bookshelf \\ 
        piano & ledge & soap dish & stool & storage bin & broom & headphones & laundry basket & tv stand \\ 
        suitcase & light switch & fire extinguisher & refrigerator & coffee maker & guitar case & toilet seat cover dispenser & shower door & closet rod \\ 
        recycling bin & closet door & ball & window & dishwasher & dustpan & bathroom stall door & curtain & plant \\ 
        laundry basket & shower floor & hat & file cabinet & machine & hair dryer & speaker & folded chair & counter \\ 
        clothes dryer & projector screen & water cooler & chair & mat & water bottle & keyboard piano & suitcase & bench \\ 
        seat & divider & mouse & wall & windowsill & handicap bar & cushion & hair dryer & ceiling \\ 
        storage bin & closet wall & scale & plant & bulletin board & purse & table & mini fridge & piano \\ 
        coffee maker & bathroom stall door & power outlet & coffee table & fireplace & vent & nightstand & dumbbell & closet \\ 
        dishwasher & stair rail & decoration & stairs & mini fridge & shower floor & bathroom vanity & oven & cabinet \\ 
        bar & bathroom cabinet & sign & armchair & water cooler & water pitcher & laptop & luggage & cup \\ 
        toaster & closet rod & projector & cabinet & shower door & bowl & shower wall & bar & laundry hamper \\ 
        ironing board & structure & vacuum cleaner & bathroom vanity & pillar & paper bag & desk & pipe & light switch \\ 
        fireplace & coat rack & candle & bathroom stall & ledge & alarm clock & computer tower & bathroom stall & cd case \\ 
        kitchen counter & storage organizer & plunger & mirror & furniture & music stand & soap dispenser & blinds & backpack \\ 
        toilet paper dispenser & ~ & stuffed animal & blackboard & cart & laundry detergent & container & toilet paper dispenser & windowsill \\ 
        mini fridge & ~ & headphones & trash can & decoration & dumbbell & bicycle & coffee table & box \\ 
        tray & ~ & broom & stair rail & closet door & tube & light & dish rack & book \\ 
        toaster oven & ~ & guitar case & box & vacuum cleaner & cd case & clothes & guitar & mailbox \\ 
        toilet seat cover dispenser & ~ & hair dryer & towel & dish rack & closet rod & machine & seat & sofa chair \\ 
        furniture & ~ & water bottle & door & range hood & coffee kettle & furniture & clock & shower curtain \\ 
        cart & ~ & purse & clothes & projector screen & shower head & stair rail & alarm clock & bulletin board \\ 
        storage container & ~ & vent & whiteboard & divider & keyboard piano & toilet paper holder & board & crate \\ 
        tissue box & ~ & water pitcher & bed & bathroom counter & case of water bottles & floor & file cabinet & tube \\ 
        crate & ~ & bowl & floor & laundry hamper & coat rack & bucket & ceiling light & window \\ 
        dish rack & ~ & paper bag & bathtub & bathroom stall door & folded chair & stool & ladder & power outlet \\ 
        range hood & ~ & alarm clock & desk & ceiling light & fire alarm & door & paper towel dispenser & power strip \\ 
        dustpan & ~ & laundry detergent & wardrobe & trash bin & power strip & sign & shower floor & bathtub \\ 
        handicap bar & ~ & object & clothes dryer & bathroom cabinet & calendar & recycling bin & stuffed animal & column \\ 
        mailbox & ~ & ceiling light & radiator & structure & poster & shower & water cooler & fire alarm \\ 
        music stand & ~ & dumbbell & shelf & storage organizer & luggage & jacket & coffee kettle & storage container \\ 
        bathroom counter & ~ & tube & ~ & potted plant & ~ & bottle & kitchen counter & ~ \\ 
        bathroom vanity & ~ & cd case & ~ & mattress & ~ & ~ & ~ & ~ \\ 
        laundry hamper & ~ & coffee kettle & ~ & ~ & ~ & ~ & ~ & ~ \\ 
        trash bin & ~ & shower head & ~ & ~ & ~ & ~ & ~ & ~ \\ 
        keyboard piano & ~ & case of water bottles & ~ & ~ & ~ & ~ & ~ & ~ \\ 
        folded chair & ~ & fire alarm & ~ & ~ & ~ & ~ & ~ & ~ \\ 
        luggage & ~ & power strip & ~ & ~ & ~ & ~ & ~ & ~ \\ 
        mattress & ~ & calendar & ~ & ~ & ~ & ~ & ~ & ~ \\ 
        ~ & ~ & poster & ~ & ~ & ~ & ~ & ~ & ~ \\ 
        ~ & ~ & potted plant & ~ & ~ & ~ & ~ & ~ & ~ \\ 
    \bottomrule
    \end{tabular}
    }
    \label{tab:split_fullview}
\end{table*}
\section{Incremental Scenarios Phases}
\label{app:split_fullview}
\Cref{tab:split_fullview} presents the task splits for each proposed scenario introduced in Section 4.3 using the ScanNet200 dataset. The three scenarios, \fsplit, \ssplit, and \rsplit, are each divided into three tasks: Task 1, Task 2, and Task 3. Notably, the order of classes in these tasks is random.

\section{Qualitative Results}
\label{app:qual_results}
\begin{figure*}[!htb]
  \centering
  \includegraphics[width=\linewidth]{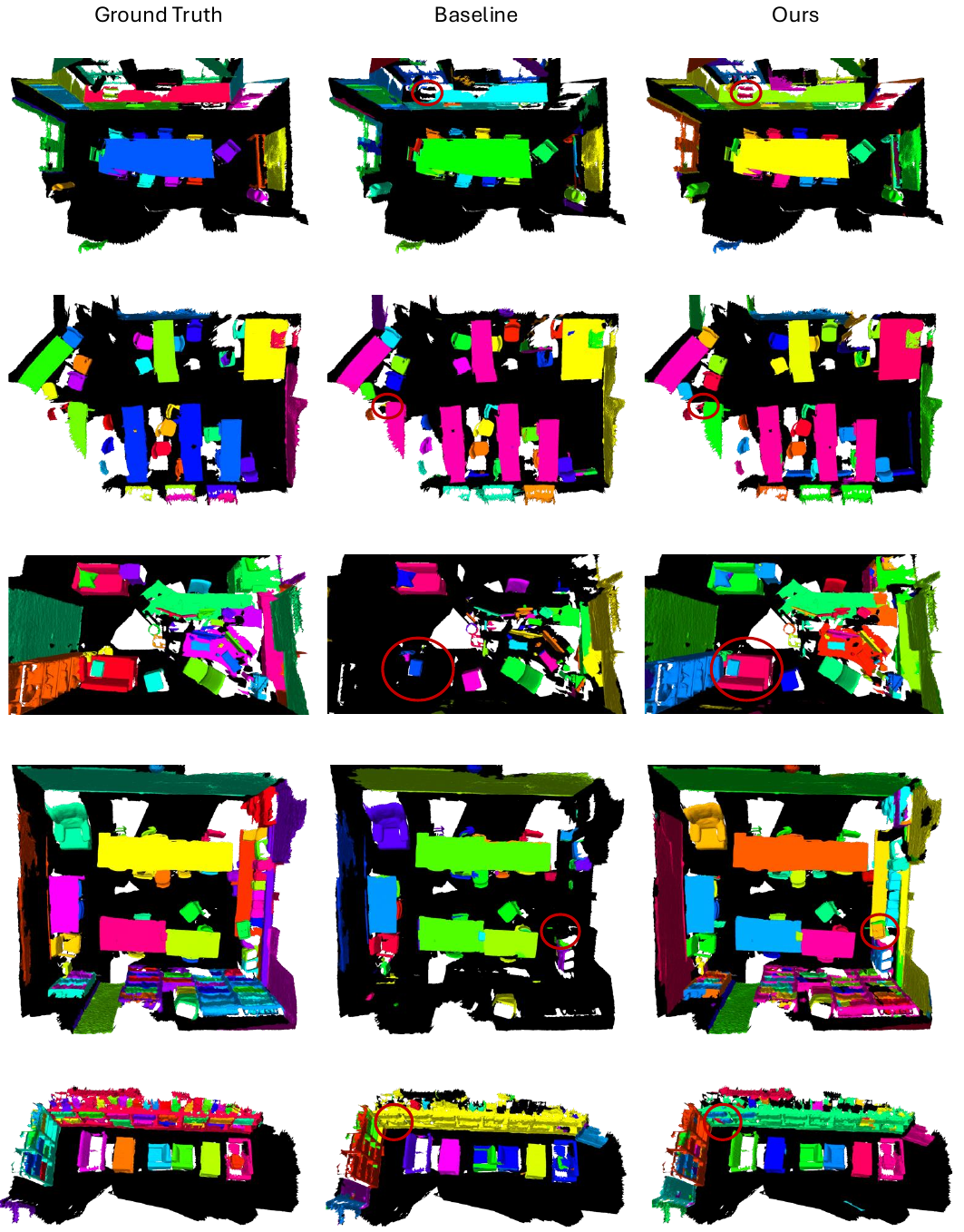}
  \caption{Qualitative comparison of ground truth, the baseline method, and our proposed framework on the \fsplit~ evaluation after learning all tasks.}
  \label{fig:fsplit_qual}
\end{figure*}

In this section, we present a qualitative comparison of the proposed framework with the baseline method. \Cref{fig:fsplit_qual} illustrates the results on the \fsplit~evaluation after learning all tasks, comparing the performance of the baseline method and our proposed approach. As shown in the figure, our method demonstrates superior instance segmentation performance compared to the baseline. For example, in row 1, the baseline method fails to segment the \texttt{sink}, while in row 3, the \texttt{sofa} instance is missed. Overall, our framework consistently outperforms the baseline, with several missed instances by the baseline highlighted in red circles.

\begin{figure*}[!htb]
  \centering
  \includegraphics[width=\linewidth]{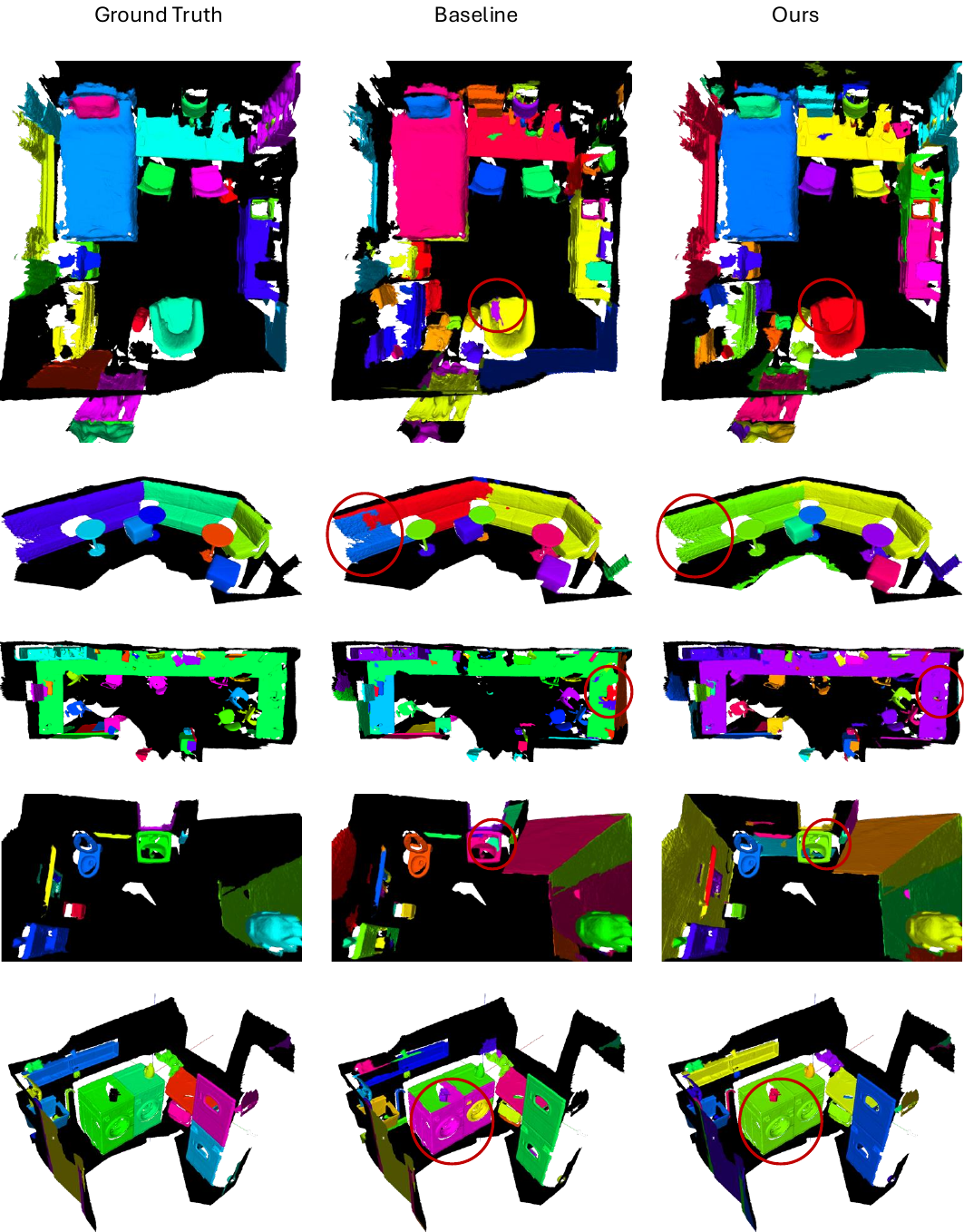}
  \caption{Qualitative comparison of ground truth, the baseline method, and our proposed framework on the \ssplit~ evaluation after learning all tasks.}
  \label{fig:ssplit_qual}
\end{figure*}

In \Cref{fig:ssplit_qual}, we present the results on \ssplit, highlighting instances where the baseline method underperforms, marked with red circles. For example, in row 2, the baseline method incorrectly identifies the same sofa as separate instances. Similarly, in row 5, the washing machine is segmented into two instances by the baseline. In contrast, the proposed method delivers results that closely align with the ground truth, demonstrating its superior performance

\begin{figure*}[!htb]
  \centering
  \includegraphics[width=\linewidth]{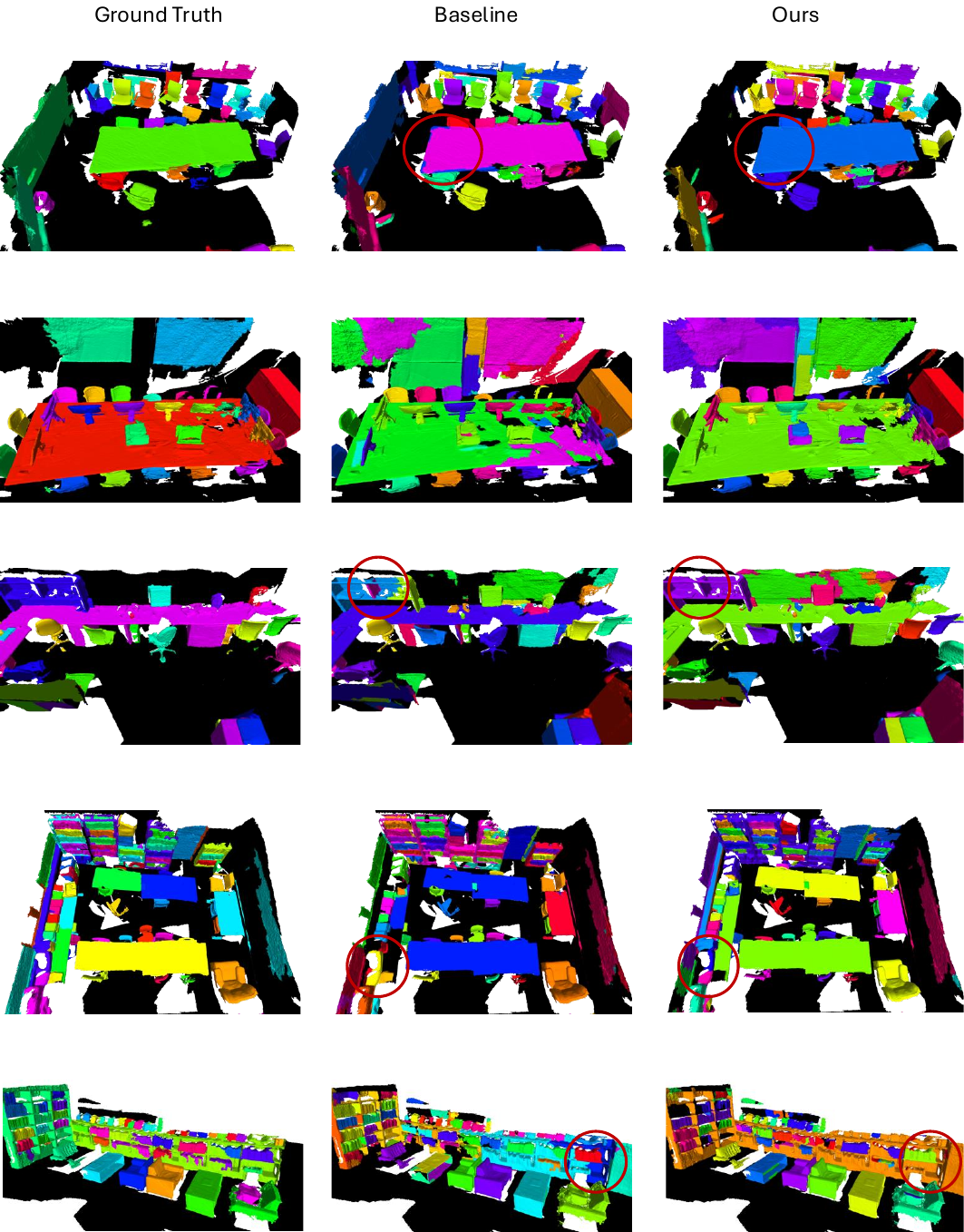}
  \caption{Qualitative comparison of ground truth, the baseline method, and our proposed framework on the \rsplit~ evaluation after learning all tasks.}
  \label{fig:rsplit_qual}
\end{figure*}

Similarly, \Cref{fig:rsplit_qual} highlights the results on \rsplit, where classes are encountered in random order. The comparison emphasizes the advantages of our method, as highlighted by red circles. The baseline method often misses instances or splits a single instance into multiple parts. In contrast, our approach consistently produces results that are closely aligned with the ground truth, further underscoring its effectiveness.

\end{document}